# 3D Masked Modelling Advances Lesion Classification in Axial T2w Prostate MRI

Alvaro Fernandez-Quilez*[1,2], Christoffer Gabrielsen Andersen[2], Trygve Eftestøl[2], Svein Reidar Kjosavik[3], and Ketil Oppedal[2]

[1]SMIL, Department of Radiology, Stavanger University Hospital, Stavanger, Norway.
[2]Department of Electrical Engineering and Computer Science, University of Stavanger, Stavanger, Norway.
[3]General Practice and Care Coordination Research Group, Stavanger University Hospital, Stavanger, Norway.

## Abstract

*Masked Image Modelling (MIM) has been shown to be an efficient self-supervised learning (SSL) pre-training paradigm when paired with transformer architectures and in the presence of a large amount of unlabelled natural images. The combination of the difficulties in accessing and obtaining large amounts of labeled data and the availability of unlabelled data in the medical imaging domain makes MIM an interesting approach to advance deep learning (DL) applications based on 3D medical imaging data. Nevertheless, SSL and, in particular, MIM applications with medical imaging data are rather scarce and there is still uncertainty around the potential of such a learning paradigm in the medical domain. We study MIM in the context of Prostate Cancer (PCa) lesion classification with T2 weighted (T2w) axial magnetic resonance imaging (MRI) data. In particular, we explore the effect of using MIM when coupled with convolutional neural networks (CNNs) under different conditions such as different masking strategies, obtaining better results in terms of AUC than other pre-training strategies like ImageNet weight initialization.* [1]

## 1 Introduction

The deep learning (DL) field has quickly progressed during the past years and its convergence with medical imaging is rapidly enabling the development of tools that can support health practitioners by automatizing tasks that otherwise would be carried out by human experts [1, 2]. In particular, 3D medical images such as MRI hold a tremendous potential to help in the management and diagnostic pathway of diseases such as PCa by, for instance, reducing the number of biopsies [3] or increasing the detection rate for the lesions that require treatment and further testing (clinically significant) [4]. Nevertheless, MRI review requires specialized training and expertise to achieve such results. Furthermore, its analysis can suffer from inter-reader variability and sub-optimal interpretation [5] and in that regard, DL applications can offer an alternative able to reduce or overcome the aforementioned issues by helping doctors in their daily chores. However, DL algorithms have traditionally relied on a supervised learning paradigm, requiring large amounts of annotated data. In the clinical domain, such an-notations are typically limited, expensive and time-consuming to obtain, making its access and availability a barrier for the progress of DL in the medical imaging domain [6, 2].

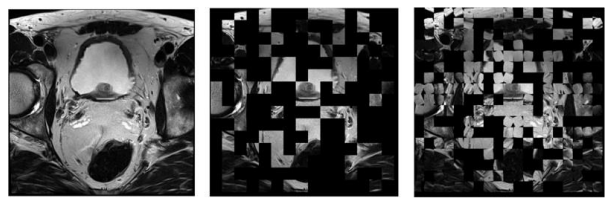

Figure 1: Example of T2w axial direction slice, static masking approach and dynamic masking (left, mid and right column, respectively).

---
*Corresponding author: alvaro.fernandez.quilez@sus.no
[1]https://github.com/cgacga/MSc_ProstateCancer

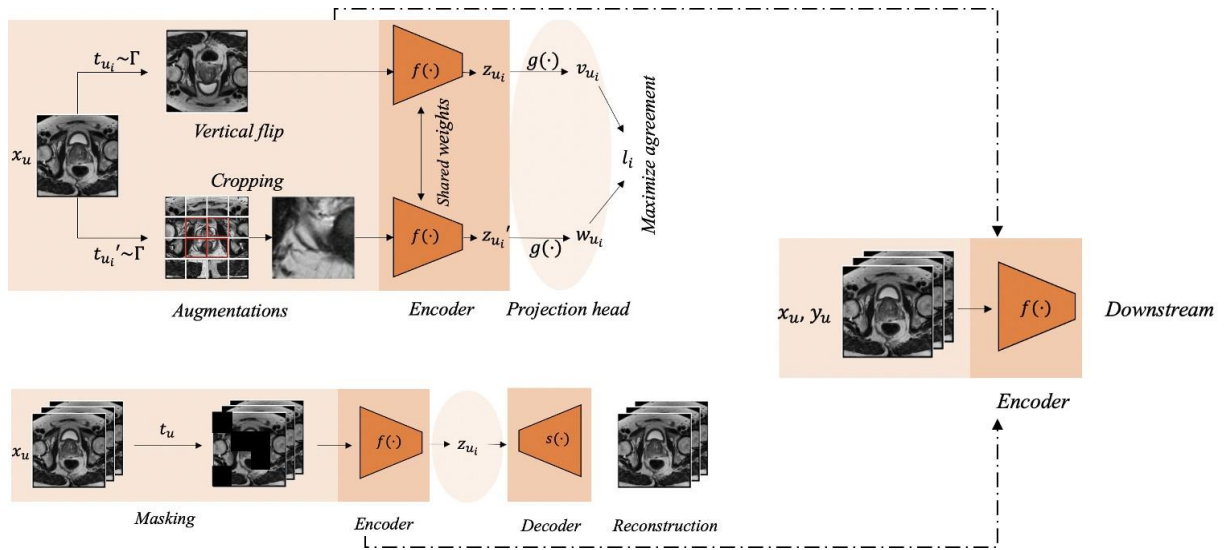

Figure 2: From left to right: Contrastive, Masked Image Modelling (MIM) and fully supervised approaches (or downstream evaluation).

Current approaches to tackle data scarcity commonly focus on transfer learning (TL) with ImageNet weights. However, the results obtained with it can be sub-optimal due to the existing gap between the initialization domain (RGB natural images) and the target domain (medical images) [7]. Self-supervised learning (SSL) [8] has aroused as a viable alternative and, in particular, has shown promising results in the medical domain when in the presence of scarce amounts of labeled data [9, 10]. However, most of the SSL approaches are tailored to natural images (RGB) and 2D data, whilst their effect is rather unknown and unexplored in the case of 3D medical images.

MIM is a type of SSL approach that uses a masked (corrupted) input to predict the unmasked original signal and has shown promising results in natural language processing (NLP) [11, 12]. Furthermore, latest approaches based on MIM have also shown a good performance in vision models [13]. In spite of it, little attention has been paid to MIM for MRI and, in particular, for PCa applications. In this work, we hypothesize that by leveraging MIM we can improve the performance of PCa diagnosis measured by area under the curve (AUC) when compared to other initialization strategies such as ImagNet and random initialization. In that regard, our contributions in the work are the following:

- We explore different masking strategies for MIM in a dynamic and static way (different ratios for each patient or a fixed ratio) in order to determine the effect of it in 3D MRI data for PCa (Figure 1).

- We compare MIM by means of fine-tuning with SSL contrastive learning approaches (SimCLR), since it has been proven to learn robust features for medical down-stream tasks [9, 10] (Figure 2).

- We explore the effect of different fractions of annotated data in a linear probing and fine-tuning setting in order to test the robustness of the approach and as a proxy for real world applications in the clinical domain, where scarce annotated data is usually the norm.

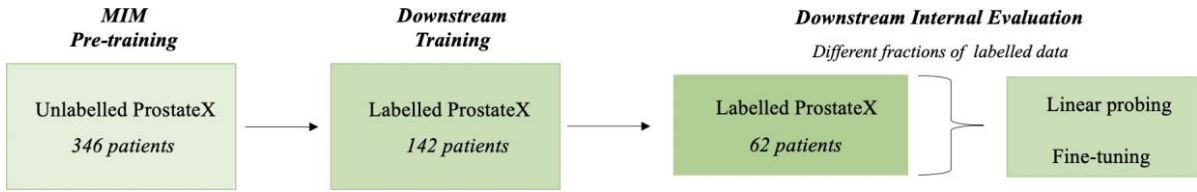

Figure 3: Masked image modelling with convolutional neural networks training and evaluation pipeline.

## 2 Data

The data used for the development of the model and internal validation is the ProstateX dataset [14]. The study is retrospective and includes different MRI sequences from which we use axial T2w due to its higher inter-plane resolution. The MRI sequences were obtained without endorectal coil with a 3.0 T field strength Siemens scanners.

The cohort included in the development of the SSL approaches consisted of 346 patients (*unlabelled* ), whilst the downstream evaluation and supervised learning approaches were evaluated on the basis of 204 labeled patients, which were the total amount available at the time the study was carried out.

The 204 labeled sequences are provided with biopsy results, which are the basis to define the *clinical significance* of the lesions for the downstream task (Figure 3). Specifically, if Gleason Score (GS) ≥the lesions are deemed to be significant and non-clinically significant otherwise. For those patients with ≥1 lesion present in the sequence, we consider the worst case scenario and label the patient according to the highest scored lesion in terms of GS score.

### 2.1 Pre-processing and data splitting

As part of the pre-processing, we re-sample all the sequences by linear interpolation to a common coordinate system with a 2D resolution of 0.5 x 0.5 $mm^2$ and 3.6 mm slice thickness, which are the predominant resolution and slice thickness among the sequences. Following, we normalize the MRI sequences intensities to a range of [0, 1] and apply outlier removal by removing the intensities of each sequence that are outside of the range of the 1st and 99th percentiles.

To train the SSL approach, we use *all the available unlabelled data* [15]. Following, we split the labeled data in 70% and 30% for training and testing sets, respectively. The training set is used to re-train the pre-trained algorithm during the downstream task whilst the testing one is used to have an estimation of the performance of the model in a real-world scenario with unseen data during training.

## 3 Approach and Evaluation

Masked image modelling (MIM) learns representations from images corrupted by masking *patches*. In particular, MIM builds on the concept of *auto-encoding*, which consists of an encoder that maps the corrupted (masked) input to a lower-dimensional representation (latent representation) and a decoder that aims to reconstruct the original input from the lower-dimensional representation *obtained from the masked input*. Following that concept, Denoising Autoencoders (DAE) [16] are a class of auto-encoders that corrupt the input signal using different corruptions such as masking pixels or rotating image patches [17].

Our masked auto-encoder (MAE) is a simple convolution-based auto-encoder that aims to recover the original 3D sequence from a partial observation, obtained by masking 3D cubes of the MRI sequence (Figure 2, bottom of left side). Specifically, we use a U-net [18] like auto-encoder architecture with a 3D VGG16-inspired [19] encoder and decoder [20] includng 3D convolutions and 3D batch normalization operations. We follow a similar training procedure as the one in [20] with an Adam optimizer and a learning rate of 0.0001. The training and inference is performed on a Tesla V100 GPU.

## 3.1 Dynamic and static approach

We split each 3D sequence into regular non-overlapping cubes. Following this, we sample a sub-set of cubes and apply a series of corruption operations (masking, rotation of 30 degrees and horizontal and vertical flipping of the cubes [20]) with a fixed size and sampling number (*static approach*) or a dynamic size and sampling for each patient (*dynamic approach*) (Figure 1). Specifically:

**Static approach.** We choose a fixed value and divide the image total size by the chosen value, such that all the cubes of the patient have the same height/width ratio. Additionally, we also fix the sub-set of cubes that will be corrupted.

**Dynamic approach** We choose a range of values and divide each patient size by a randomly chosen value in that range, such that all the patients have a different height/width ratio cubes. Additionally, we also set a corruption range, such that the sub-set of cubes that will be corrupted is different for every patient.

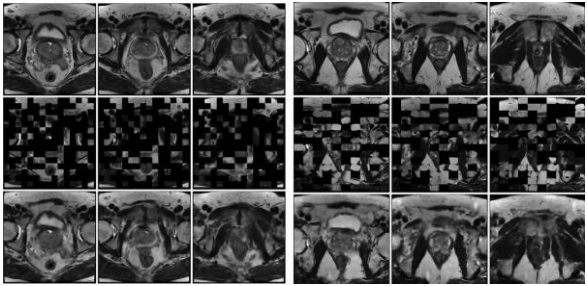

Figure 4: Reconstructions obtained with Static (three first columns) and Dynamic (last three columns) MIM approaches.

We conducted experiments to test the effect of two different approaches to the MIM pre-training strategy. Specifically, we experiment with the static and dynamic corruption approaches and different configurations of them. As shown in Table 1, the main difference lie on the fact that we apply different corruptions, sub-sampling %, cube sizes and corruption operations probabilities (that is, for a specific patient a corruption might happen to a cube with a pre-defined probability $p$). Examples of the of the effect of the approaches in selected slices of patients is shown in Figure 1 (we carry it out in 3D but for the sake of simplicity we exemplify it in 2D, at the slice level) whilst the effect of the two approaches in the reconstruction of the patient sequences when carrying out the pre-training stage, is depicted in Figure 4.

Table 1: Characteristics of the different tested MIM approaches.

| Characteristics | Approach | |
|---|---|---|
| | Static | Dynamic |
| Cube size | 32x32x16 | [9x9x2 -32x32x16] |
| Corruptions | Occlusion | Occlusion, Rotation & Flipping |
| Sub-sampling % | 60 | [60 - 90] |
| Occlusion/Others ratio % | 100 | 50 |

## 3.2 Linear probing and fine-tuning

We evaluate the effect of proposed pre-training strategy following previous work practices on unsupervised visual representation learning [13]. In particular, we evaluate the downstream task (*lesion classification*) under two different settings: *Linear probing* and *fine-tuning*. In the first case, the weights obtained from the MIM pre-training strategy are frozen and a randomly initialized linear head (*fully-connected layer*) is trained for the lesion classification task. The idea underlying such an evaluation protocol is to obtain an estimation of the quality of the learnt features and their re-usability [9]. In the other case scenario, the encoder is unfrozen and re-trained. Both procedures are carried out with different fractions of labeled data: 10, 25, 50 and 100%. That is, we sample in a stratified way the chosen % of the test set without replacement and evaluate the method with that sub-set of the test set (Figure 2).

## 3.3 Bootstrapping and statistical significance

As part of the evaluation protocol, we include a non-parametric bootstrap approach with $n = 100$ replicates from the test set to estimate the variability of the model performance. Following, we obtain the 95 % bootstrap confidence intervals (CI) and assess the significance at the $p$ = 0.05 level between the different approaches (dynamic, static and baselines) in terms of AUC. The comparison is carried out by means of Wilcoxon signed-rank [21].

Table 2: Results for MIM (dynamic approach), ImageNet and Random initialization with 100% of labeled data and linear probing evaluation approach.

| Method | AUC | Accuracy | Precision | Recall | F1 | p value |
|---|---|---|---|---|---|---|
| Random | 0.64 [0.53, 0.70] | 0.60 [0.48, 0.69] | 0.54 [0.42, 0.56] | 0.53 [0.43, 0.64] | 0.53 [0.42, 0.60] | 0.050† |
| ImageNet | 0.71 [0.62, 0.80] | 0.66 [0.56, 0.78] | 0.62 [0.46, 0.78] | 0.60 [0.47, 0.71] | 0.60 [0.46, 0.74] | 0.492 |
| MIM | **0.75** [0.69, 0.83] | **0.71** [0.63, 0.78] | **0.73** [0.53, 0.87] | **0.60** [0.51, 0.69] | **0.66** [0.52, 0.70] | - |

† Statistically significant.

All the conducted experiments include two stages: pre-training with the MIM approach and the downstream task evaluation. We use the same encoder of the architecture used for the MIM approach (VGG16) and train it with ImageNet weights and random initialization. Both approaches are considered to be the baselines of the work and compared against MIM under the different evaluation protocols described in Section 3.2. The results are presented in terms of bootstrapping (Section 3.3) and AUC, Accuracy, Precision, Recall and F1 metrics (*mean* and 95% CI). In particular, the $p$ values are obtained based on the AUC. Following, we present a sensitivity analysis on the effect of the static and dynamic approaches presented in Section 3.1 and the different corruption operations and sub-sampling amount tested in such approaches.

## 4 Results

### 4.1 Linear probing

We start by investigating if the representations learnt by MIM are of higher quality than the ones transferred through ImageNet or obtained from a random initialisation. As depicted in Figure 5, the MIM approach shows statistically significant improvements when compared to a random initialization for different % of annotated data but shows non-significant improvements when compared to ImageNet initialization. When tested with *small amounts of labeled data*, the MIM approach reaches an averaged AUC *0.66* [0.62, 0.74], compared with *0.58* [0.51, 0.68] of ImageNet and *0.51* [0.50, 0.55] of Random initialisation. These finding support the hypothesis that MIM representations are of higher quality, which is most apparent when the annotated data is scarce and when compared to a random initialization. In particular, Table 2 shows the results for all the metrics when the methods are presented with *100% of annotated data*. As the table shows, MIM outperforms the other initialization methods based on AUC results when all the original annotated data is available, showing the efficiency of using in-domain data rather than out-of-domain methods.

### 4.2 Fine-tuning

Figure 6 shows the proposed MIM method with a dynamic approach offers significant improvements when evaluated under a *fine-tuning* protocol ($p < 0.001$) and when compared to ImageNet and Random initialisation for all the different % of annotated data. In particular, the MIM improvements are particularly large for small amounts of annotated data (Random *0.51* [0.52, 0.55], ImageNet *0.58* [0.51, 0.60] and MIM *0.66* [0.62, 0.74]), showing that the MIM pre-training strategy is more efficient than its out-of-domain counterparts when in the presence of scarce amounts of annotated data.

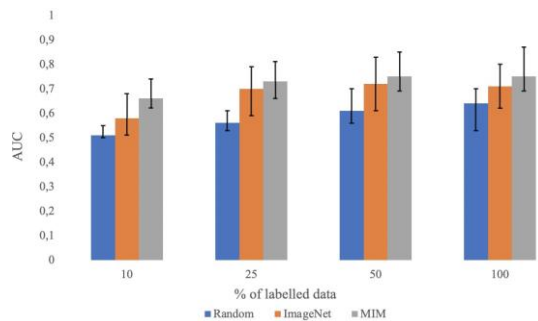

Figure 5: AUC and 95% CI intervals for different fractions of labeled data and Random, ImageNet and MIM initialisation with a dynamic approach and linear probing evaluation scheme.

Table 3: Results for MIM (dynamic approach), ImageNet and Random initialization with 100% of labeled data and fine-tuning evaluation approach.

| Method | Metrics | | | | | p value |
|---|---|---|---|---|---|---|
| | AUC | Accuracy | Precision | Recall | F1 | |
| Random | 0.64 [0.56, 0.80] | 0.67 [0.56, 0.80] | 0.52 [0.26, 0.84] | 0.53 [0.44, 0.57] | 0.52 [0.33, 0.68] | <0.001[†] |
| ImageNet | 0.77 [0.68, 0.86] | 0.73 [0.64, 0.81] | 0.72 [0.42, 1.0] | 0.55 [0.47, 0.67] | 0.62 [0.44, 0.80] | 0.568 |
| MIM | **0.80** [0.71, 0.90] | **0.78** [0.70, 0.84] | **0.75** [0.46, 0.94] | **0.58** [0.49-0.70] | **0.65** [0.48, 0.80] | - |

† Statistically significant.

Specifically, as shown in Table 3 the *dynamic MIM* approach outperforms the other initialisation methods in terms of the other metrics used in this work (Accuracy, Precision, Recall and F1) when in the presence of *100% of annotated data* and in a *fine-tuning* evaluation protocol.

## 4.3 Dynamic vs static approach

We observe non-significant differences for the AUC between the dynamic and static approach (*0.80* [0.72-0.90] vs *0.71* [0.62-0.81], p = 0.08) and larger values when it comes to accuracy, precision and recall (Figure 7). Specifically, we observe that the dynamic approach obtains better results both in a scarce amount of data setting (10% of annotated data) in terms of AUC (*0.62* [0.58-0.71] vs *0.67* [0.65-0.74] p = 0.215) and with the original amount of annotated data (*0.70* [0.58-0.81] vs *0.80* [0.71-0.90] p = 0.08).

## 4.4 Contrastive learning vs MIM

We experiment with SimCLR [9] and compare the results in a fine-tuning setting with our dynamic MIM approach in terms of AUC for 10% and 100% of annotated data. We observe that the dynamic MIM outperforms SimCLR in both settings in terms of AUC (*0.80* [0.71-0.90] vs *0.77* [0.69-0.84] *p* = 0.501 for 100% and *0.66* [0.62-0.74] vs *0.62* [0.61-0.69]) p = 0.278 for 10%), depicting its robustness in the presence of the original amount of annotated data and in an evaluation setting simulating an extreme data scarcity.

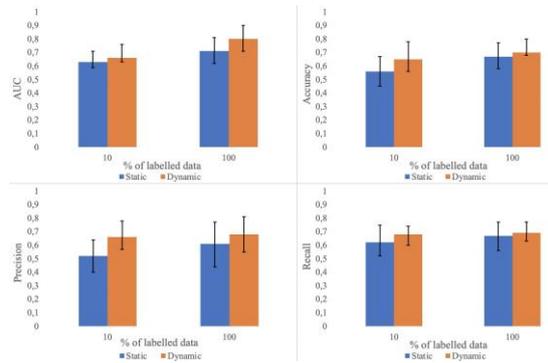

Figure 7: Results for dynamic and static MIM approaches (AUC, Accuracy, Precision and Recall, from left to right) in a fine-tuning setting and different fractions of annotated data.

## 5 Conclusions

To the best of our knowledge, this is the first contribution that shows the potential of masked modelling for MRI in a PCa context. We experimented with different masking approaches showing that a dynamic per patient corruption MIM pre-training scheme is able to obtain high quality data features for PCa lesion classification, even in the presence of highly scarce labeled data. In particular, we observe that a dynamic MIM approach outperforms random initialisation, ImageNet and SimCLR approaches, showing larger AUC when compared to standard initialisation techniques (out-of-domain) and other self-supervised in-domain ones (SimCLR).

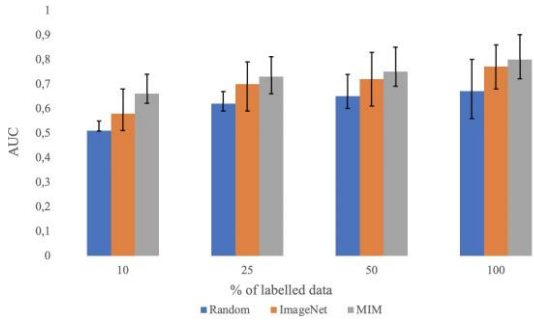

Figure 6: AUC and 95% CI intervals for different fractions of labeled data and Random, ImageNet and MIM initialisation with a dynamic approach and fine-tuning evaluation scheme.

# References

[1] Mohammad Hossein Jarrahi. Artificial intelligence and the future of work: Human-ai symbiosis in organizational decision making. *Business horizons*, 61(4):577–586, 2018. doi: 10.1016/j.bushor.2018.03.007.

[2] Alvaro Fernandez-Quilez. Deep learning in radiology: ethics of data and on the value of algorithm transparency, interpretability and explainability. *AI and Ethics*, pages 1–9, 2022. doi: 10.1007/s43681-022-00161-9.

[3] Martha MC Elwenspoek, Athena L Sheppard, Matthew DF McInnes, Samuel WD Merriel, Edward WJ Rowe, Richard J Bryant, Jenny L Donovan, and Penny Whiting. Comparison of multiparametric magnetic resonance imaging and targeted biopsy with systematic biopsy alone for the diagnosis of prostate cancer: a systematic review and meta-analysis. *JAMA network open*, 2(8):e198427–e198427, 2019. doi: 10.1001/jamanetworkopen.2019.8427.

[4] Matthias Röthke, AG Anastasiadis, M Lichy, M Werner, P Wagner, S Kruck, Claus D Claussen, A Stenzl, HP Schlemmer, and D Schilling. Mri-guided prostate biopsy detects clinically significant cancer: analysis of a cohort of 100 patients after previous negative trus biopsy. *World journal of urology*, 30(2):213–218, 2012. doi: 10.1007/s00345-011-0675-2.

[5] Andrew B Rosenkrantz, Luke A Ginocchio, Daniel Cornfeld, Adam T Froemming, Rajan T Gupta, Baris Turkbey, Antonio C Westphalen, James S Babb, and Daniel J Margolis. Interobserver reproducibility of the pi-rads version 2 lexicon: a multicenter study of six experienced prostate radiologists. *Radiology*, 280(3):793, 2016. doi: 10.1148/radiol.2016152542.

[6] Andre Esteva, Katherine Chou, Serena Yeung, Nikhil Naik, Ali Madani, Ali Mottaghi, Yun Liu, Eric Topol, Jeff Dean, and Richard Socher. Deep learning-enabled medical computer vision. *NPJ digital medicine*, 4(1):1–9, 2021. doi: 10.1038/s41746-020-00376-2.

[7] Maithra Raghu, Chiyuan Zhang, Jon Kleinberg, and Samy Bengio. Transfusion: Understanding transfer learning for medical imaging. *Advances in neural information processing systems*, 32, 2019.

[8] Longlong Jing and Yingli Tian. Self-supervised visual feature learning with deep neural networks: A survey. *IEEE transactions on pattern analysis and machine intelligence*, 43(11):4037–4058, 2020. doi: 10.1109/TPAMI.2020.2992393.

[9] Alvaro Fernandez-Quilez, Trygve Eftestøl, Svein Reidar Kjosavik, Morten Goodwin, and Ketil Oppedal. Contrasting axial t2w mri for prostate cancer triage: A self-supervised learning approach. In *2022 IEEE 19th International Symposium on Biomedical Imaging (ISBI)*, pages 1–5, 2022. doi: 10.1109/ISBI52829.2022.9761573.

[10] Shekoofeh Azizi, Basil Mustafa, Fiona Ryan, Zachary Beaver, Jan Freyberg, Jonathan Deaton, Aaron Loh, Alan Karthikesalingam, Simon Kornblith, Ting Chen, et al. Big self-supervised models advance medical image classification. In *Proceedings of the IEEE/CVF International Conference on Computer Vision*, pages 3478–3488, 2021.

[11] Jacob Devlin, Ming-Wei Chang, and Kenton Lee. Kristina, toutanova. *Bert: Pre-training of deep bidirectional, transformers for language understanding. In, NAACL*, 2(3), 2019. doi: 10.18653/v1/N19-1423.

[12] Tom Brown, Benjamin Mann, Nick Ryder, Melanie Subbiah, Jared D Kaplan, Prafulla Dhariwal, Arvind Neelakantan, Pranav Shyam, Girish Sastry, Amanda Askell, et al.

Language models are few-shot learners. *Advances in neural information processing systems*, 33:1877–1901, 2020.

[13] Kaiming He, Xinlei Chen, Saining Xie, Yanghao Li, Piotr Dollár, and Ross Girshick. Masked autoencoders are scalable vision learners. In *Proceedings of the IEEE/CVF Conference on Computer Vision and Pattern Recognition*, pages 16000–16009, 2022.

[14] Geert Litjens, Oscar Debats, Jelle Barentsz, Nico Karssemeijer, and Henkjan Huisman. Computer-aided detection of prostate cancer in mri. *IEEE transactions on medical imaging*, 33(5):1083–1092, 2014. doi: 10.1109/TMI.2014.2303821.

[15] Hari Sowrirajan, Jingbo Yang, Andrew Y Ng, and Pranav Rajpurkar. Moco pretraining improves representation and transferability of chest x-ray models. In *Medical Imaging with Deep Learning*, pages 728–744. PMLR, 2021.

[16] Pascal Vincent, Hugo Larochelle, Isabelle Lajoie, Yoshua Bengio, Pierre-Antoine Manzagol, and Léon Bottou. Stacked denoising autoencoders: Learning useful representations in a deep network with a local denoising criterion. *Journal of machine learning research*, 11(12), 2010.

[17] Pascal Vincent, Hugo Larochelle, Yoshua Bengio, and Pierre-Antoine Manzagol. Extracting and composing robust features with denoising autoencoders. In *Proceedings of the 25th international conference on Machine learning*, pages 1096–1103, 2008.

[18] Olaf Ronneberger, Philipp Fischer, and Thomas Brox. U-net: Convolutional networks for biomedical image segmentation. In *International Conference on Medical image computing and computer-assisted intervention*, pages 234–241. Springer, 2015. doi: 10.1007/978-3-319-24574-4_28.

[19] Karen Simonyan and Andrew Zisserman. Very deep convolutional networks for large-scale image recognition. *arXiv preprint arXiv:1409.1556*, 2014.

[20] Alvaro Fernandez-Quilez, Trygve Eftestøl, Svein Reidar Kjosavik, and Ketil Oppedal. Learning to triage by learning to reconstruct: a generative self-supervised approach for prostate cancer based on axial t2w mri. In *Medical Imaging 2022: Computer-Aided Diagnosis*, volume 12033, pages 460–466. SPIE, 2022. doi: 10.1117/12.2610623.

[21] Robert F Woolson. Wilcoxon signed-rank test. *Wiley encyclopedia of clinical trials*, pages 1–3, 2007. doi: 10.1002/9780471462422.eoct979.